\documentclass[sigconf]{acmart} 

\usepackage{subcaption}
\usepackage{multirow}
\usepackage{makecell}
\newcolumntype{P}[1]{>{\raggedright\arraybackslash}p{#1}}

\AtBeginDocument{%
  }

\acmSubmissionID{v1rtp0903}

\copyrightyear{2026}
\acmYear{2026}
\setcopyright{cc}
\setcctype{by}
\acmConference[KDD '26]{Proceedings of the 32nd ACM SIGKDD Conference on Knowledge Discovery and Data Mining V.1}{August 09--13, 2026}{Jeju Island, Republic of Korea}
\acmBooktitle{Proceedings of the 32nd ACM SIGKDD Conference on Knowledge Discovery and Data Mining V.1 (KDD '26), August 09--13, 2026, Jeju Island, Republic of Korea}
\acmPrice{}
\acmDOI{10.1145/3770854.3780299}
\acmISBN{979-8-4007-2258-5/2026/08}

\begin{document}

\title{Instruction-based Time Series Editing}

\author{Jiaxing Qiu}
\affiliation{%
  \institution{University of Virginia}
  \city{Charlottesville}
  \state{VA}
  \country{USA}
}
\email{jq2uw@virginia.edu}

\author{Dongliang Guo}
\affiliation{%
  \institution{University of Virginia}
  \city{Charlottesville}
  \state{VA}
  \country{USA}
}
\email{ktm8eh@virginia.edu}

\author{Brynne Sullivan}
\affiliation{%
  \institution{University of Virginia}
  \city{Charlottesville}
  \state{VA}
  \country{USA}
}
\email{bsa4m@uvahealth.org}

\author{Teague R. Henry}
\affiliation{%
  \institution{University of Virginia}
  \city{Charlottesville}
  \state{VA}
  \country{USA}
}
\email{ycp6wm@virginia.edu}

\author{Thomas Hartvigsen}
\affiliation{%
  \institution{University of Virginia}
  \city{Charlottesville}
  \state{VA}
  \country{USA}
}
\email{hartvigsen@virginia.edu}

\renewcommand{\shortauthors}{Jiaxing Qiu, Dongliang Guo, Brynne Sullivan, Teague R. Henry, \& Thomas Hartvigsen}

\begin{abstract}

In time series editing, we aim to modify some properties of a given time series without altering others. For example, when analyzing a hospital patient’s blood pressure, we may add a sudden early drop and observe how it impacts their future, while preserving other conditions. Existing diffusion-based editors rely on rigid, predefined attribute vectors as conditions and produce all-or-nothing edits through sampling. This attribute- and sampling-based approach limits flexibility in condition format and lacks customizable control over editing strength. To overcome these limitations, we introduce Instruction-based Time Series Editing, where users specify intended edits using natural language. This allows users to express a wider range of edits in a more accessible format. We then introduce InstructTime, the first instruction-based time series editor. InstructTime takes in time series and instructions, embeds them into a shared multi-modal representation space, then decodes their embeddings to generate edited time series. By learning a structured multi-modal representation space, we can easily interpolate between embeddings to achieve varying degrees of edit. To handle local and global edits together, we propose multi-resolution encoders. In our experiments, we use synthetic and real datasets and find that InstructTime is a state-of-the-art time series editor: InstructTime achieves high-quality edits with controllable strength, can generalize to unseen instructions, and can be easily adapted to unseen conditions through few-shot learning. 
 
\end{abstract}



\maketitle
\newcommand\kddavailabilityurl{https://doi.org/10.5281/zenodo.18098681}
\ifdefempty{\kddavailabilityurl}{}{
\begingroup\small\noindent\raggedright\textbf{Resource Availability:}\\
The source code of this paper has been made publicly available at \url{https://github.com/JiaxingQiu/InstructTime}.
\endgroup
}

\section{Introduction}\label{sec:introduction}

\begin{figure}
    \centering
    \includegraphics[width=\linewidth]{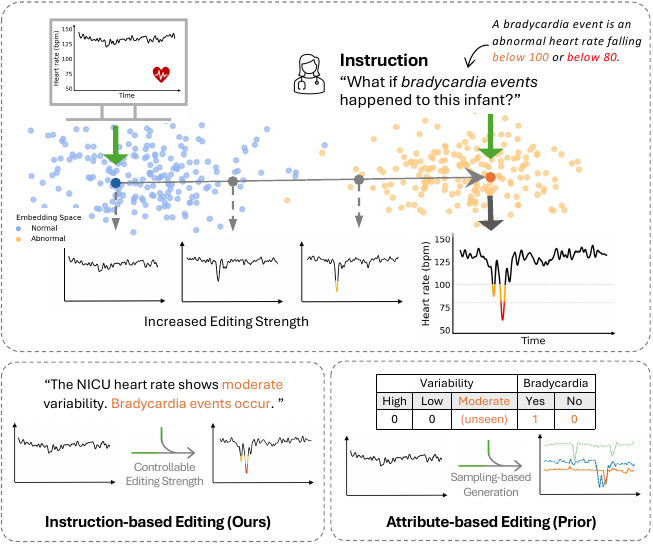}
    \caption{\small \textit{Instruction-based time series editing} modifies a given time series using natural language instructions (e.g., adding abnormal events to a normal heart rate). Our approach enables controllable editing strength and generalizes to unseen instructions. In contrast, existing attribute- and sampling-based methods edit time series with uncontrolled strength based on predefined attribute vectors.}
    \label{fig:figure1}
\end{figure}

Many recent works have focused on time series generation \cite{TimeGAN,Sommers2024,Wen2023,Rasul2021,CSDI,TimeDiT,Diff-MTS, esteban2017real,TimeVAE,CGAN,CGAN,yoon2019time} as a way to synthesize realistic data for time series analysis tasks such as classification \cite{wen2020time, kulevome2024effective, villegas2024data}, forecasting \cite{semenoglou2023data, chen2023fraug, bandara2021improving}, and counterfactual reasoning \cite{bandara2021improving, li2025controllable, bunne2024build, lee2024clinical}.
A recent extension is Time Series Editing (TSE), which enables fine-grained manipulation of an input time series by directly modifying it to match desired conditions while preserving the characteristics of the original data \cite{jing2024towards, narasimhan2024time, yu2025unlock}. 
%
TSE differs from the time series generation task in that it modifies an input time series toward a target condition rather than generating a completely new time series from scratch \cite{jing2024towards}.
For example, as illustrated in Figure \ref{fig:figure1}, a user may wonder ``given a normal heart rate time series, what would happen if my patient's heart rate rapidly falls below 80 beats per minute?'' This describes \textit{bradycardia}, a dangerous situation for hospitalized infants \cite{ambalavanan2023cardiorespiratory}. A successful time series editor would take in the normal heart rate and new condition (bradycardia happened), then generate a new time series showing the likely locations, severity, and shapes of such abnormalities if they were to occur. 
However, no existing methods can handle natural language edits.

State-of-the-art TSE methods focus on attribute-level control and use conditional diffusion \cite{jing2024towards, narasimhan2024time}. 
However, these methods have two limitations.
Firstly, they represent attributes as continuous or categorical feature vectors. These attributes require handcrafted feature engineering, and constrain all future edits to capture only these attributes, and limits users to this hardcoded interaction with the editor.
Secondly, diffusion-based methods perform all-or-nothing edits by sampling diverse outputs, offering no control over editing strength. In contrast, allowing edits with varying condition influence enables finer user control and supports insight and hypothesis generation—e.g., gradually adding abnormality to a normal heart rate from mild to severe.

We propose a new time series editing task: \textbf{Instruction-based Time Series Editing}. 
An editor takes in a time series and a text instruction specifying target conditions and generates a modified time series that reflects the target conditions.
Instead of hard-coding attributes, users can directly guide time series editing using natural language instructions, which flexibly integrate structured and unstructured, qualitative and quantitative conditions into a unified semantic representation.
In real-world settings, text paired with time series—such as nurse notes or patient narratives—can capture nuanced, individualized conditions under which the time series is generated. Instruction-based time series editing addresses this setting by enabling editors to condition on heterogeneous free-text inputs, where condition factors vary widely across instances.

Instruction-based time series editing poses three key challenges.
First, an editor should handle multiple conditions described in a single natural language instruction and fuse them with a time series exhibiting multi-resolution patterns—some conditions affect global patterns such as trends and seasonality, while others lead to localized changes such as abrupt mean shifts or abnormalities. 
Secondly, since conditions are expressed discretely in natural language \cite{studdert2005did}, fine-grained control over editing strength is especially necessary. For example, rather than editing a heart rate time series from ``normal'' to ``abnormal'', the user may progressively add abnormality at mild, moderate, and severe levels during editing. To allow such interpolated editing, the editor must encode the varying degrees to which an instruction can influence a time series.
Lastly, the editor should leverage the richness of natural language and generalize to new expressions of known conditions or to unseen conditions, based on their semantic relationships to observed ones.

To overcome these challenges, we introduce \textbf{InstructTime}, the first instruction-based time series editing method.
InstructTime maps time series and text instructions into a shared unit-length hypersphere via contrastive learning \cite{radford2021learning}, and decodes linearly-interpolated embeddings of a time series and its target instruction into new time series that satisfy the instructed conditions to varying degrees.
We evaluate InstructTime on both synthetic and real-world datasets. Under instruction-based settings, our model achieves the state-of-the-art performance, while enabling interpolated editing and generalizability to unseen instructions with few-shot tuning. 
We also evaluate a categorical attribute version of InstructTime to fairly compare with recent non-instruction-based time series editors, and find that it achieves higher fidelity and comparable performance to state-of-the-art methods.

In summary, our contributions are as follows: 
\begin{enumerate}
    \item We introduce instruction-based time series editing, a novel problem setting that lets users edit time series using natural language instructions. 
    \item We propose InstructTime, the first instruction-based time series editor, which enables multi-resolution time series editing via multi-condition instructions, enabling high-fidelity editing with controllable strength. 
    \item Our experiments show InstructTime successfully enables instruction-based editing, achieves state-of-the-art performance, and generalizes to unseen instructions.
\end{enumerate}

\vspace{-10pt}
\section{Related Works}\label{sec:related_works}

\begin{figure*}
    \centering
    \includegraphics[width=\textwidth]{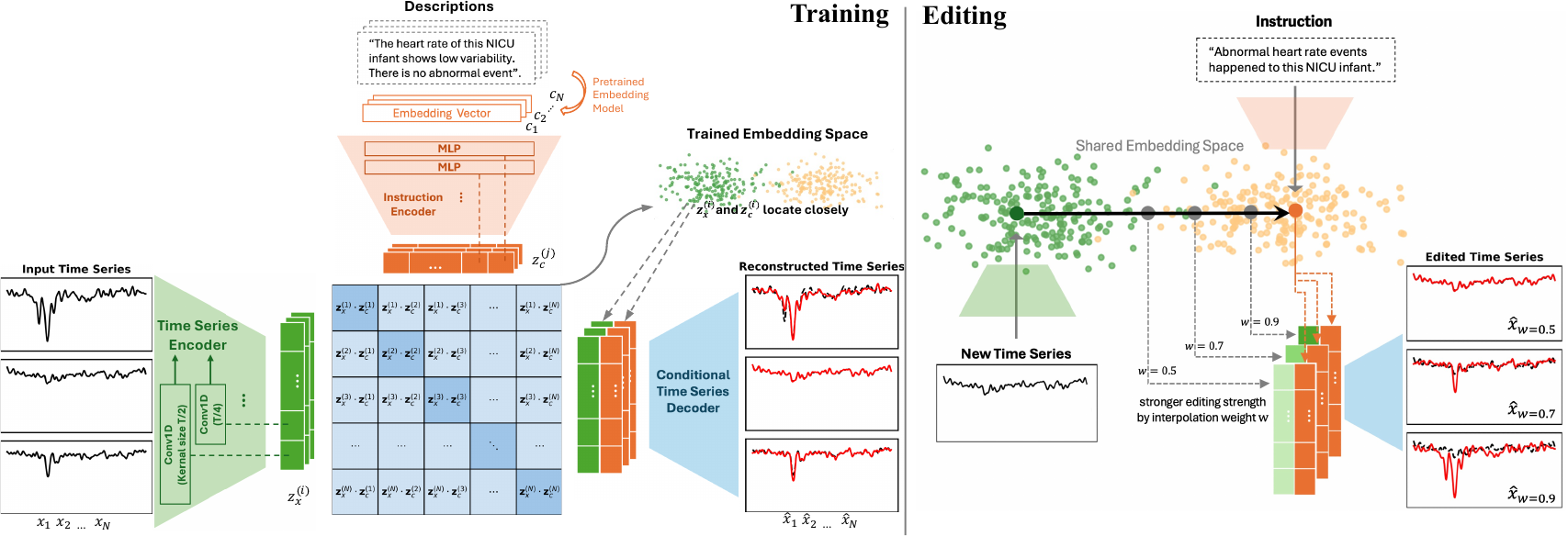}
    \caption{\small InstructTime architecture, training, and interpolated editing.  InstructTime includes a multi-resolution time series encoder, an instruction encoder, and a time series conditional decoder.  During training (left), time series–description pairs are encoded into latent representations $z_x$ and $z_y$ in a shared hyperspherical space, and the decoder reconstructs the input time series from both. During editing (right), we encode a time series and an instruction onto the shared embedding space, then decode interpolated embeddings between them to generate edits at varying editing strength.}
    \label{fig:architecture}
\end{figure*}

\noindent\textbf{Time-Series Generation.} 
The generation of time-series data has progressed through advancements in both unconditional and conditional generative models. 
\textit{Unconditional generation} focuses on producing time-series data without explicit constraints. Early approaches employed generative adversarial networks \cite{TimeGAN,Sommers2024,Wen2023} and diffusion models \cite{Rasul2021,CSDI,TimeDiT} to synthesize realistic sequences. 
Unconditional models lack the ability to guide the generation toward desired outcomes.
\textit{Conditional generation} provides finer control over the generated time series by conditioning on external signals. Early efforts explored global attribute control via extensions of GANs and VAEs~\cite{TimeVAE,TimeGAN,CGAN}. 
More recent developments introduced hierarchical frameworks~\cite{Torres2021} and attention-based architectures~\cite{Liu2024}, enabling more granular control. 
Diffusion-based conditional models have emerged~\cite{narasimhan2024time,coletta2023constrained,Tian2024,Hamdouche2023}, offering improved controllability and sample diversity. 
Several recent works have used language prompts to generate synthetic time series 
\cite{huang2025timedp, li2025bridge, grigoraș2024synthetic}. Nonetheless, these methods focus on generating new time series from scratch without editing an input, thus are not directly comparable to the editing task we consider.

\vspace{5pt}
\noindent\textbf{Time Series Editing.}
Time series editing differs from the time series generation task in that it modifies \textit{an input time series} to satisfy target conditions while preserving the underlying structure of the original data as much as possible. It enables more fine-grained manipulation while retaining the specific patterns of the input. 
In contrast, a conditional time series generation model takes only a condition as input and generate a new time series from scratch. It samples from the target conditional distribution without the need to preserve any properties of an existing time series.
Recent approaches build upon conditional diffusion models with enhancements such as multi-resolution encoding~\cite{jing2024towards, narasimhan2024time}.
TEdit~\cite{jing2024towards} extends TimeWeaver~\cite{narasimhan2024time}, a conditional diffusion-based generation method, into an editing framework by introducing a generation process that uses a denoising diffusion sampler (e.g., the denoising diffusion implicit model~\cite{song2020denoising}) along with a multi-resolution architecture. It takes as input a triplet of an input time series, its original attribute vector, and the target attribute vector, and generates an edited time series where the attributes that differ from the original are modified, while the remaining ones are preserved.
These methods require conditions to be categorical or continuous attribute vectors, limiting generalizability to new conditions during inference. Moreover, while diffusion-based editors can produce diverse outputs through stochastic sampling, they lack mechanisms to explicitly regulate the degree of editing. In this study, we take a step further by allowing users to edit time series using natural language instructions and controllable editing strength.

\vspace{5pt}
\noindent\textbf{Contrastive Learning for Time Series–Language Fusion.}
Contrastive learning has emerged as a powerful paradigm for aligning heterogeneous modalities in a shared embedding space. Pioneered by CLIP~\cite{radford2021learning} for image–language alignment, contrastive methods have since been extended to time series for tasks such as retrieval, classification, and forecasting. Recent works~\cite{chen2025trace,hoxha2025self,cai2025semi,HAO2023MICOS} show that time series can be effectively aligned with textual descriptions or labels via contrastive objectives, enabling cross-modal retrieval and classification.  
To learn better time series representations, several studies~\cite{zheng2024parametric,liu2023timesurl,Liu2023Self} use synthetic descriptions or structured metadata to supervise alignment, while others~\cite{zheng2024simple,woo2022cost,HU202386} align future trends with language cues to improve temporal generalization.  
While these efforts demonstrate the effectiveness of contrastive learning in fusing time series and language, they are largely limited to static latent representation alignment and are only evaluated on retrieval, classification, or forecasting tasks. 
A detailed comparison on problem definitions of the related tasks is provided in Table~\ref{problem_definition}.

\begin{table}[t]
\centering
\caption{Differences in problem definitions of related tasks.}
\small
\resizebox{0.48\textwidth}{!}{%
\begin{tabular}{@{}P{2.0cm}P{0.2cm}P{1.2cm}P{0.7cm}P{3cm}P{2cm}@{}}
\toprule
 & \multicolumn{2}{c}{\textbf{Input}} & \textbf{Output} & \textbf{Objectives} & \textbf{Methods} \\ 
\cmidrule(lr){2-3}
\textbf{Task} & \textbf{\footnotesize TS} & \textbf{\footnotesize Condition} & \textbf{\footnotesize TS} &  &  \\ 
\midrule
Conditional Time Series Generation & \text{\large \textcolor{red}{$\times$}} & \text{\large $\checkmark$} & \text{\large $\checkmark$} & \footnotesize (1) generate a new time series from scratch, following target distribution. & \cite{TimeVAE,TimeGAN,CGAN,coletta2023constrained,Tian2024,Hamdouche2023, huang2025timedp, li2025bridge, grigoraș2024synthetic} \\
Time Series--Text Fusion & \text{\large $\checkmark$} & \text{\large $\checkmark$} & \text{\large \textcolor{red}{$\times$}}& \footnotesize (1) representation learning. & \cite{chen2025trace,hoxha2025self,cai2025semi,HAO2023MICOS,zheng2024parametric,liu2023timesurl,Liu2023Self, zheng2024simple,woo2022cost,HU202386} 
\\
Time Series Editing & \text{\large $\checkmark$} & \text{\large $\checkmark$} \newline (Attributes) & \text{\large $\checkmark$} & \footnotesize (1) \textit{editability}: edit a given time series towards target conditions; & \footnotesize TEdit~\cite{jing2024towards}, TimeWeaver~\cite{jing2024towards, narasimhan2024time} \\
 & \text{\large $\checkmark$} & \text{\large $\checkmark$}  \newline (Instruction) & \text{\large $\checkmark$} & (2) \footnotesize \textit{preservability}: preserve other conditions of the input time series. & \footnotesize \textbf{InstructTime (Ours)} \\ 
\bottomrule
\end{tabular}\label{problem_definition}
}
\end{table}

\section{Instruction-based Time Series Editing}\label{sec:methods}

\subsection{Problem Formulation}\label{problem_def}

Let $\mathbf{x} \in \mathbb{R}^{T}$ be a time series with $T$ timesteps, and let $\mathbf{c} = [c_1, c_2, \ldots, c_L]$ be a text instruction consisting of $L$ tokens, describing the target condition of the edited time series. The target condition refers to the realization of a set of $\mathrm{K}$ attributes on the target time series, $\Tilde{\mathbf{c}} = \left\{ \Tilde{c}_i \right\}_{i=1}^{\mathrm{K}}$, some of which may differ from those of the input time series, while others remain the same.

\textbf{Definition 1 (Instruction-based Time Series Editing).} 
\textit{Given an input time series $\mathbf{x}$ and a natural language instruction $\mathbf{c}$, the instruction-based time series editing task is to build a function $f_{\theta}$ that modifies $\mathbf{x}$ into a new time series $\hat{\mathbf{x}} = f_{\theta}(\mathbf{x}, \mathbf{c})$, such that $\hat{\mathbf{x}}$ reflects the target condition described in the instruction $\mathbf{c}$. }

Compared to prior work~\cite{jing2024towards}, we use a text instruction $\mathbf{c}$ instead of a feature vector $\Tilde{\mathbf{c}}$ to guide the editing. We also relax prior problem definition by \textbf{not requiring the original condition of $\mathbf{x}$ as an additional input}, because the model should be expected to infer the original condition directly from $\mathbf{x}$ without relying on external manual specification. 
An effective instruction‑based editor should also parse multiple conditions described in the instruction and modify the subset that differ from the input’s conditions while preserving those that align—thereby achieving both editability and preservability~\cite{jing2024towards}—in the edited time series.

In this section, the terms ``description'' and ``instruction'' are both used to denote textual inputs. We use ``description'' to refer to the text paired with each time series that is fused with the time series during training, while “instruction” indicates the user-specified input at inference time that guides the time series editing process.

\subsection{InstructTime}\label{architecture}

We propose InstructTime, the first instruction-based time series editor, leveraging CLIP-based contrastive learning and a novel interpolated editing procedure for controllable editing strength.
As illustrated in Figure~\ref{fig:architecture} (left), InstructTime combines a multi-resolution time series encoder, an instruction encoder generalizable to diverse semantic expressions of conditions, and a time series decoder that generates time series by modeling both intra- and inter-modality relationships via a self-attention mechanism.

\vspace{5pt}
\noindent\textbf{Multi-resolution time series encoder $\mathcal{E}_{\phi}$.} The time series encoder $\mathcal{E}_{\phi}$ transforms a time series $\mathbf{x}$ with $T$ timesteps into a latent embedding $\mathbf{z}_x$ in the $D$-dimensional unit-length hypersphere,
\[
\mathbf{z}_x = \mathcal{E}_{\phi}(\mathbf{x}) \quad \text{s.t. } \| \mathbf{z}_x \|_2 = 1.
\]
A key challenge in encoding a time series is capturing multi-resolution features of the input $\mathbf{x}$, as meaningful patterns can occur at different temporal scales—ranging from global trends  (e.g., accelerating heart rate) to local fluctuations (e.g., abrupt drops in heart rate). To effectively embed information across multiple resolutions, we use a convolutional neural network (CNN)-based encoder $\mathcal{E}_{\phi}$ with varying kernel sizes (Figure~\ref{fig:architecture}). Such CNN-based multi-resolution design has proven effective in both time series analysis tasks \cite{cirstea2022towards, wang2023mianet, jing2024towards, mohammadi2024deep, qian2020dynamic} and multi-scale image segmentation and classification \cite{yu2023convolutions, he2016accurate, yang2021cmf, lin2023clip}. $\mathcal{E}_{\phi}$ consists of $k$ parallel 1D CNNs, each with a kernel size proportional to $T$. Varying kernel sizes capture patterns at different temporal resolutions, with larger kernels capturing global patterns such as trends and seasonality, while smaller kernels are more effective at detecting local variations such as abrupt shifts or outliers. 
Each CNN outputs a $d$-dimensional vector $\mathbf{z}_{x_k}$, capturing a latent representation of the time series at a distinct temporal resolution. The $k$ vectors are then concatenated to form the final $D$-dimensional time series embedding, $\mathbf{z}_x = [\mathbf{z}_{x_1} \,\|\, \mathbf{z}_{x_2} \,\|\, \cdots \,\|\, \mathbf{z}_{x_k}]$, where $D = k \cdot d$.
Lastly, $\mathbf{z}_x$ is normalized to have unit length, constraining the embedding space to a hypersphere and preventing uncontrolled distances between time series in an unbounded space \cite{davidson2018hyperspherical}.

Regarding the architecture, each CNN consists of a 1D convolutional layer with stride = 1 and zero-padding set to half of the kernel size to preserve temporal length. Each CNN is followed by a ReLU activation, batch normalization, a max-pooling layer with stride = 2, and a linear projection to the $d$-dimensional embedding space. The design choices for $k$ and the fractions that determine kernel sizes proportional to $T$ depend on the datasets and experimental settings, and are therefore described in the implementation details of the experimental setup (Section~\ref{eval_setup}).

\vspace{5pt}
\noindent\textbf{Instruction encoder $\mathcal{E}_{\theta}$.}
The text encoder $\mathcal{E}_{\theta}$ encodes the instruction $\mathbf{c}$ into an embedding $\mathbf{z}_c$, which also lies in the $D$-dimensional unit-length hypersphere,
\vspace{-5pt}
\[ 
\mathbf{z}_c = \mathcal{E}_{\theta}(\mathbf{c}) \quad \text{s.t. } \| \mathbf{z}_c \|_2 = 1.
\]
To efficiently compress an instruction into a semantic embedding, we first leverage a pretrained text embedding model (frozen parameters) to encode the instruction $\mathbf{c}$ into a numeric vector.
Such vector captures rich semantic meaning and paraphrastic similarity of the multiple conditions described in the instruction \cite{reimers-2019-sentence-bert, jiang2025explainable}.
The vector is then processed by $k$ parallel multi-layer perceptrons (MLPs), each predicting a $d$-dimensional vector $\mathbf{z}_{c_k}$. These vectors are concatenated to form the final $D$-dimensional instruction embedding, $\mathbf{z}_c = [\mathbf{z}_{c_1} \,\|\, \mathbf{z}_{c_2} \,\|\, \cdots \,\|\, \mathbf{z}_{c_k}]$, where $D = k \cdot d$.
This chunked representation enables resolution-wise semantic alignment—where the $r^{\text{th}}$ segment of the instruction embedding aligns with the $r^{\text{th}}$ segment of the time series embedding ($r = {1, \ldots, k}$)—thereby linking each segment’s semantic representation to its corresponding temporal resolution during both contrastive fusion and decoding. 
A detailed explanation of how the contrastive learning encourages resolution-wise semantic alignment is provided in Appendix~\ref{sup:proof_alignment}.
Lastly, $\mathbf{z}_c$ is also normalized to have unit length to ensure it resides in the same bounded embedding space as $\mathbf{z}_x$.

Regarding the architecture, each MLP consists of LayerNorm, a two-layer feedforward network with GELU activations and a linear projection to the $d$-dimensional embedding space.

\vspace{5pt}
\noindent\textbf{Conditional time series decoder $\Psi$.}
Given a time series embedding $\mathbf{z}_x$ and an instruction embedding $\mathbf{z}_c$, the goal of the decoder $\Psi$ is to generate an edited time series $\hat{\mathbf{x}}$ of length $T$ that reflects the target condition encoded in $\mathbf{z}_c$ while preserving the unique characteristics of the input time series encoded in $\mathbf{z}_x$,
\vspace{-5pt}
\[
\hat{\mathbf{x}} = \Psi(\mathbf{z}_x, \mathbf{z}_c).
\]
Our time series decoder $\Psi$ leverages the self-attention mechanism from the Transformer architecture \cite{vaswani2017attention}, allowing it to model complex dependencies between the time series and instruction embeddings. It takes the time series and instruction embeddings as a token sequence of length 2, applies positional encoding, and processes them through eight attention-based blocks, each consisting of multi-head self-attention, feedforward layers, and residual connections. The self-attention mechanism allows each token to attend to both itself and other tokens. This enables the model to capture relationships within the time series embedding (e.g., between multi-resolution representations) and within the instruction embedding (e.g., between multiple conditions). It also models the relationships between the two modalities with respect to the mappings between the multi-resolution representations of time series and the multiple semantic conditions.
Lastly, the output hidden state corresponding to the time series token is passed through a linear head to generate a time series $\hat{\mathrm{x}}$ of the same length $T$ as the input time series $\mathrm{x}$.

\vspace{5pt}
\noindent\textbf{Training.} The goal of training is to learn a shared embedding space where time series and instruction embeddings align semantically, while accurately reconstructing the original time series from both modalities.
InstructTime can be trained on datasets containing pairs of time series and corresponding descriptions $\left\{ (\mathrm{x}_i, \mathrm{c}_i) \right\}_{i=1}^{N}$, where $N$ denotes the sample size, by minimizing the following loss function:
\[
\mathcal{L} = \mathcal{L}_{\text{contrast}}(\mathrm{z}_{\mathrm{x}}, \mathrm{z}_{\mathrm{c}}) + \alpha \cdot \mathcal{L}_{\text{recon}}(\mathrm{x}, \hat{\mathrm{x}});
\]
\[
\resizebox{0.45\textwidth}{!}{$
\mathcal{L}_{\text{contrast}} = \frac{1}{2N} \sum_{i=1}^{N} [
    -\log\left(
        \frac{\exp( \mathrm{z}_{\mathrm{x}}^{(i)} \cdot \mathrm{z}_{\mathrm{c}}^{(i)} )}
             {\sum_{j=1}^{N} \exp( \mathrm{z}_{\mathrm{x}}^{(i)} \cdot \mathrm{z}_{\mathrm{c}}^{(j)} )}
    \right)
    -\log\left(
        \frac{\exp( \mathrm{z}_{\mathrm{c}}^{(i)} \cdot \mathrm{z}_{\mathrm{x}}^{(i)} )}
             {\sum_{j=1}^{N} \exp( \mathrm{z}_{\mathrm{c}}^{(i)} \cdot \mathrm{z}_{\mathrm{x}}^{(j)} )}
    \right)
], 
$}
\]
\[\small
\mathcal{L}_{\text{recon}} = \frac{1}{N} \sum_{i=1}^{N} \left\| \mathrm{x}_i - \hat{\mathrm{x}}_i \right\|_2^2.
\]
The contrastive loss $\mathcal{L}_{\text{contrast}}$ is the symmetric InfoNCE loss used in CLIP \cite{radford2021learning} and the reconstruction loss $\mathcal{L}_{\text{recon}}$ is the mean squared error (MSE) between $\mathrm{x}$ and its reconstruction $\hat{\mathrm{x}}$.
We train the model in two steps:
(1) we first minimize $\mathcal{L}_{\text{contrast}}$ to update the parameters of $\mathcal{E}_{\phi}$ and $\mathcal{E}_{\theta}$, encouraging paired embeddings $\mathrm{z}_{\mathrm{x}}$ and $\mathrm{z}_{\mathrm{c}}$ to move closer on the unit-length $D$-dimensional hypersphere;  
(2) we then jointly minimize $\mathcal{L}_{\text{contrast}}$ and $\mathcal{L}_{\text{recon}}$ through the overall loss $\mathcal{L}$ to update the parameters of both encoders and the decoder $\Psi$. This allows the model to reconstruct the time series under the constraint that the time series and instruction embeddings reside closely in a shared embedding space.
In step (2), to prevent the model from de-prioritizing contrastive learning, the balancing weight $\alpha$ controls the unit contribution of the reconstruction loss relative to the contrastive loss. 
The training configurations in our experiments on various datasets are described in Section~\ref{eval_setup} \textit{training details}.

\subsection{Interpolated Editing Procedure}\label{generation}

To control editing strength at inference, we use an interpolated editing procedure (Figure~\ref{fig:architecture}, right).
It is motivated by our embedding geometry and encoder–decoder design: the shared encoder places similar time series close together and distinct ones farther apart in the unit-length subspace, creating geometric continuity that bridges discrete textual conditions. For instance, time series with large positive slopes (“strongly upward”) lie near those with small positive (“weakly upward”) and small negative (“weakly downward”) slopes, forming a smooth semantic continuum from upward to downward. Interpolation between input time series and instruction embeddings thus traces a smooth transition between their conditions. 
Similar interpolation procedures have been used in text \cite{bowman2016generating}, image \cite{wang2023interpolating, ramesh2022hierarchical}, and motion \cite{tevet2022motionclip} generation tasks.

Given a time series and an instruction, their embeddings $\mathrm{z}_x$ and $\mathrm{z}_c$ are projected onto a shared unit-length $D$-dimensional hypersphere.
We linearly interpolate between $\mathrm{z}_x$ and $\mathrm{z}_c$ with weight $w \in [0,1]$ and normalize the embedding to obtain the interpolated embedding $\mathrm{z}_w$.
The decoder $\Psi$ then maps $\mathrm{z}_w$ to a time series that gradually transitions toward the target condition as $w$ increases.
\[
\mathrm{z}_w = \frac{(1-w)\,\mathrm{z}_x + w\,\mathrm{z}_c}
{\left\|(1-w)\,\mathrm{z}_x + w\,\mathrm{z}_c\right\|}, \qquad
\hat{\mathrm{x}}_{w} = \Psi(\mathrm{z}_w, \mathrm{z}_c).
\]
Larger $w$ result in stronger editing, making the target condition more prominent in the edited time series, whereas smaller $w$ produce time series that more closely resemble the input time series while moderately reflecting the target condition. For example, editing a downward trend toward an upward one, increasing $w$ from 0 to 1 shifts the trend from downward to flat, then to upward.  
When $w = 0$, the decoder reconstructs the input: $\hat{\mathrm{x}}_{w=0} = \Psi(\mathrm{z}_{x}, \mathrm{z}_\mathrm{c})$. When $w = 1$, the decoder generates a new time series solely from text instruction: $\hat{\mathrm{x}}_{w=1} = \Psi(\mathrm{z}_\mathrm{c}, \mathrm{z}_\mathrm{c})$. This means InstructTime supports both time series editing and conditional generation.
We also experimented with spherical linear interpolation for unit-norm hypersphere geometry. We find no notable performance difference—due to tightly clustered embeddings—and greater numerical stability for normalized linear interpolation for nearly parallel embeddings.

\subsection{Few-shot Tuning on Unseen Conditions}
Editing time series via natural language enables generalization to rich semantic expressions, including (1) novel phrasings of known conditions and (2) instructions for entirely unseen conditions.
In (1), a pretrained text encoder helps $\mathcal{E}_{\theta}$ map new expressions close to seen ones in embedding space, enabling zero-shot generalization.
In (2), however, it's a strong assumption that unseen instructions align with their (also unseen) time series in embedding space, and that $\Psi$ can decode them correctly in a zero-shot manner.
To enable data-efficient tuning on conditions unseen during training, we also introduce a few-shot tuning procedure. Assuming users can provide a small number of time series–instruction pairs from the unseen condition, we adopt a two-step approach: first, edit each new time series toward seen conditions using varying interpolation weights (e.g., 0.1 to 0.9) to create synthetic examples paired with seen instructions; then, tune model parameters on the combined set of synthetic and unseen pairs by minimizing the loss $\mathcal{L}$.

\vspace{-10pt}
\section{Experiments}\label{sec:experiments}

In this section, we evaluate InstructTime on three key capabilities—editing quality, interpolated editing, and generalizability to unseen instructions. We use both a synthetic dataset where time series are associated with both global and local attributes, and two real-world datasets. Performance is assessed across five metrics measuring both editability and preservability. Lastly, we conduct a sensitivity analysis on three key components: the balancing weight in the loss function controlled by $\gamma$, the number of resolutions $k$, and the choice of pretrained text embedding model.

\vspace{-10pt}
\subsection{Experimental Setup}\label{eval_setup}

\noindent\textbf{Datasets.} To evaluate InstructTime, we introduce one new synthetic dataset and use two real-world datasets, one from prior time series editing papers, and one new task.
For the \textbf{Synthetic} dataset, we generate 18,000 time series samples of length 200, each being a mix of five multi-resolution attributes: trend direction (3 levels: upward, downward, none), trend type (2 levels: linear, quadratic), seasonality (2 levels: present or absent), abrupt mean shift (3 levels: upward, downward, none), and noise level (2 levels: high or low). Trend, seasonality, and noise level represent global patterns, while abrupt mean shifts reflect local patterns. 
We next use \textbf{Air Quality}, which combines two public datasets \cite{godahewa2020kddcup, chen2017beijing} used in prior works \cite{jing2024towards, narasimhan2024time}, resulting in 3,684 particulate matter (PM2.5) time series from Beijing and London spanning 2013 to 2018. We process the combined datasets following the same procedure as \cite{jing2024towards}. The resulting time series each have a length of 168 and are influenced by two attributes: location (2 levels) and season (4 levels). 
Third, we draw from a real neonatal intensive care unit dataset to create a new time series editing dataset called \textbf{NICU Heart Rate}, which contains 36,679 heart rate time series of length 300 (10-minute records sampled every 2 seconds) from 2,780 NICU infants between 2012 and 2016 \cite{nicudataset, sullivan2024comparing}. Each time series is labeled with two clinically meaningful attributes: heart rate variability (2 levels: high or low), and bradycardia events (2 levels: whether or not any heart rate drops below 100 bpm) \cite{ambalavanan2023cardiorespiratory, qiu2024highly}.
To prevent large-scale values during training, both the Air Quality and NICU Heart Rate datasets are normalized by subtracting the global mean and dividing by the global standard deviation, calculated from all values across all time series.
All datasets are randomly split into 70\% for training, 20\% for validation, and 10\% for held-out evaluation.
Details on datasets can be found in Appendix~\ref{sup:datasets}.

\vspace{5pt}
\noindent\textbf{Compared methods.} We compare InstructTime with two state-of-the-art methods—TimeWeaver \cite{narasimhan2024time} and TEdit \cite{jing2024towards}—both of which are implemented with the sampling-based editing procedure introduced in~\cite{jing2024towards}. 
Since we relax the input requirement from requiring a triplet—comprising a time series, its original attributes, and target attributes—to only taking a time series and target attributes, we adopt the denoising diffusion probabilistic model sampler implemented in~\cite{jing2024towards}, which does \textbf{not} require the original attributes as input. To use the same computing resource, modifications on configuration are verified to achieve same performance to the original configuration in the attribute-based editing setting.

\noindent\textbf{Implementation details.} For InstructTime, the time series encoder uses $k=8$ parallel CNNs with kernel sizes proportional to the time series length $T$, using fractions ${1, \frac{2}{3}, \frac{1}{2}, \frac{1}{3}, \frac{1}{4}, \frac{1}{6}, \frac{1}{8}, \frac{1}{10}}$. Larger fractions (${1, \frac{2}{3}, \frac{1}{2}}$) are used to capture global properties such as trend and seasonality, while smaller fractions are intended to encode localized patterns such as abrupt shifts in mean and local variability.
For the synthetic data where $T=200$, such design results in kernel sizes of $\{200,133,100,66,50,33,25,20\}$ after truncation toward zero. For the air quality data where $T=168$, the resulting kernel sizes are $\{168,112,84,56,42,28,21,16\}$. For the NICU data where $T=300$, the resulting kernel sizes are $\{300,200,150,100,75,50,37,30\}$.
Text instructions are encoded using the pretrained SentenceTransformer model \textit{paraphrase-mpnet-base-v2}~\cite{reimers-2019-sentence-bert}, which provides effective embedding of rich semantic meaning from multiple sentences and is efficient to compute. The resulting vectors are then processed by $k=8$ parallel MLPs to generate instruction embeddings $\mathbf{z}_{\mathrm{c}}$.
The embeddings $\mathbf{z}_{\mathrm{x}}$ and $\mathbf{z}_{\mathrm{c}}$ have dimension $D = 768$. When training InstructTime, we jointly optimize a contrastive loss and a reconstruction loss, with their relative contributions to the total loss controlled by a balancing weight $\alpha$. In our experiments, $\alpha$ is set so that the unit contribution of the reconstruction loss is $10^{-\gamma}$ relative to the contrastive loss, with $\gamma$ set to 1 by default.

\vspace{5pt}
\noindent\textbf{Training details.} InstructTime is trained on all datasets using a single NVIDIA V100 GPU, requiring 10 GB, 5 GB, and 20 GB of GPU memory for the synthetic, air quality, and NICU datasets, respectively. The corresponding training times are within 0.5, 0.5, and 1 hour. We use an initial learning rate of \(1\times10^{-4}\) with a Reduce-on-Plateau learning rate scheduler \cite{loshchilov2017decoupled} that decreases the learning rate by a factor of 0.9 when the validation loss plateaus. The scheduler has a patience of 50 epochs and a cool-down period of 20 epochs after each reduction. Optimization is performed using AdamW optimizer \cite{loshchilov2017decoupled} with a weight decay of \(1\times10^{-4}\). Early stopping is applied based on validation loss.

\vspace{5pt}
\noindent\textbf{Instruction-based vs. attribute-based editing.}
We compare methods in both instruction-based and attribute-based editing settings.
In the \textbf{instruction-based} setting, each level of each time series attribute is described using a unique natural language template (e.g., “This is air quality in [Beijing / London]”). The full set of templates is provided in Appendix~\ref{sup:instruction}. Then sentences of all attributes are concatenated into a paragraph-length instruction.
To evaluate generalizability to \textbf{unseen expressions}, the description of each attribute level is augmented into 50 different expressions using GPT-4o\cite{hurst2024gpt}, 30\% of which are left out for evaluation. Details on the prompting are provided in Appendix~\ref{sup:instruction}.
Existing methods encode a vector of categorical or continuous attributes into an attribute embedding and process it through MLP layers in the conditional diffusion model. 
We replace the attribute embedding with the instruction embedding from the pretrained model, ensuring identical input to the MLP layers in our instruction encoder.
In the \textbf{attribute-based} setting, we input the vector of categorical attributes to InstructTime's instruction encoder, matching the input format used by existing methods.

\begin{table*}[htbp]
\centering
\resizebox{\textwidth}{!}{%
\begin{tabular}{llccccccccccc}
\toprule
 &  & \multicolumn{5}{c}{\textbf{Synthetic}} & \multicolumn{3}{c}{\textbf{Air quality}} & \multicolumn{3}{c}{\textbf{NICU heart rate}} \\
\cmidrule(lr){3-7}\cmidrule(lr){8-10}\cmidrule(lr){11-13}
 &   & \footnotesize $\Delta$DTW~$\downarrow$& \footnotesize RaTS ↑& \footnotesize |RaTS| ↓& \footnotesize MSE ↓& \footnotesize MAE ↓& \footnotesize $\Delta$DTW~$\downarrow$& \footnotesize RaTS ↑& \footnotesize |RaTS| ↓& \footnotesize $\Delta$DTW~$\downarrow$& \footnotesize RaTS ↑& \footnotesize |RaTS| ↓\\
\midrule
\multirow[c]{3}{*}{\makecell{Instruction-based}} & Time Weaver & -11.05 & 5.93 & 0.63 & 3.50 & 1.54 & -0.97 & 0.17 & 0.66 & -0.15 & 0.13 & 1.03 \\
 & TEdit & -9.66 & 5.99 & 0.86 & 3.45 & 1.52 & -0.90 & 0.12 & 0.65 & -2.41 & 0.15 & 1.03 \\
 & InstructTime & \textbf{-14.62} & \textbf{6.87} & \textbf{0.02} & \textbf{2.26} & \textbf{1.18} & \textbf{-1.84} & \textbf{0.51} & \textbf{0.62} & \textbf{-4.38} & \textbf{0.18} & \textbf{0.42} \\
\midrule
\multirow[c]{3}{*}{\makecell{Attribute-based}} & Time Weaver & -12.19 & 6.65 & 0.15 & 3.58 & 1.54 & -1.26 & 0.31 & 0.73 & -1.55 & 0.15 & 1.40 \\
 & TEdit & -10.66 & \textbf{6.98} & \textbf{0.08} & 3.96 & 1.61 & \textbf{-1.71} & 0.30 & 0.72 & -1.96 & 0.18 & \textbf{1.38} \\
 & InstructTime & \textbf{-13.62} & 6.77 & 0.20 & \textbf{2.38} & \textbf{1.25} & -1.62 & \textbf{0.43} & \textbf{0.70} & \textbf{-4.94} & \textbf{0.26} & 1.41 \\
\bottomrule
\end{tabular}
}
\caption{Comparison of model editing quality across datasets under instruction-based vs. attribute-based editing settings. Scores are averaged across attributes. InstructTime results are reported at $w = 0.9$.}
\label{tab:comparison_table}
\end{table*}

\vspace{5pt}
\noindent\textbf{Evaluation metrics.} An editor should demonstrate both editability and preservability \cite{jing2024towards}. This means identifying the set of $\mathrm{K}$ attributes ($\Tilde{\mathbf{c}} = \left\{ \Tilde{\mathrm{c}}_i \right\}_{i=1}^{\mathrm{K}}$) described in an instruction, and then generating a time series $\hat{\mathrm{x}}$ where the subset of attributes to edit ($\Tilde{\mathbf{c}}_{edit}$) is modified and the subset to preserve ($\Tilde{\mathbf{c}}_{prsv}$) is retained. 
We adopt two established metrics to evaluate the editability and preservability: Log Ratio of Target-to-Source (RaTS) \cite{jing2024towards} and Dynamic Time Warping distance (DTW) \cite{sakoe2003dynamic}.  
(1) RaTS measures the extent to which an output time series $\hat{\mathrm{x}}$ is more likely to exhibit an attribute $\Tilde{\mathrm{c}}$ than the input time series $\mathrm{x}$:
\vspace{-10pt}
\[
\text{RaTS}(\hat{\mathrm{x}}, \mathrm{x}, \Tilde{\mathrm{c}}) = \log \left( \frac{p(\Tilde{\mathrm{c}} \mid \hat{\mathrm{x}})}{p(\Tilde{\mathrm{c}} \mid \mathrm{x})} \right),
\]
where $p(\Tilde{\mathrm{c}} | \mathrm{x})$ is the probability of $\hat{\mathrm{x}}$ being classified to the correct level of attribute $\Tilde{\mathrm{c}}$. 
For each attribute, we compute $p(\Tilde{\mathrm{c}} | \mathrm{x})$ using a multi-class classifier that appends a softmax prediction head to the architecture of time series encoder $\mathcal{E}_{\phi}$, trained end-to-end on the held-out dataset.
Higher $\text{RaTS}(\hat{\mathrm{x}}, \mathrm{x}, \Tilde{\mathrm{c}}_{\text{edit}})$ indicates better editability. 
The absolute value $|\text{RaTS}(\hat{\mathrm{x}}, \mathrm{x}, \Tilde{\mathrm{c}}_{\text{prsv}})|$ measures how well the output $\hat{\mathrm{x}}$ preserves the remaining attributes $\Tilde{\mathrm{c}}_{\text{prsv}}$ of the input $\mathrm{x}$; values closer to zero indicate better preservability~\cite{jing2024towards}.
Note that we use an end-to-end trained classifier to calculate $p(\Tilde{\mathrm{c}} | \mathrm{x})$ for each attribute, rather than using the retrieval softmax probability computed from the shared embedding space as done by~\cite{jing2024towards}, because our generation process used the shared embedding space, and using retrieval softmax probability may favor our edits. Our results remain consistent across runs, while uncertainty in metrics can occur due to the sampling procedure in prior methods.  

(2) $\text{DTW}$ measures the similarity between two time series by finding an optimal alignment that minimizes the cumulative point-wise difference \cite{sakoe2003dynamic}. We use $\Delta\text{DTW}$, defined below, to measure how much closer $\hat{\mathrm{x}}$ is to the data observed with target attributes $\Tilde{\mathbf{c}}$ (denoted as $\mathcal{D}_{\Tilde{\mathbf{c}}}$) compared to $\mathrm{x}$:
\[
\Delta \mathrm{DTW}(\hat{\mathrm{x}},\mathrm{x}, \mathcal{D}_{\Tilde{\mathbf{c}}}) = \mathrm{median}_{\mathrm{x}_{\Tilde{\mathbf{c}}} \in \mathcal{D}_{\Tilde{\mathbf{c}}}} \left[ \mathrm{DTW}(\mathrm{x}, \mathrm{x}_{\Tilde{\mathbf{c}}}) - \mathrm{DTW}(\hat{\mathrm{x}}, \mathrm{x}_{\Tilde{\mathbf{c}}}) \right].
\]
A lower negative $\Delta \mathrm{DTW}$ suggests better editability because the edited time series is closer to the real data with the target attributes.
Following \cite{jing2024towards}, for synthetic data with ground truth, we also report MSE and MAE to evaluate the point-wise closeness between the ground truth $x_{\text{gt}}$ and the generated time series $\hat{\mathrm{x}}$. 
\begin{figure}
    \centering
    \includegraphics[width=\linewidth]{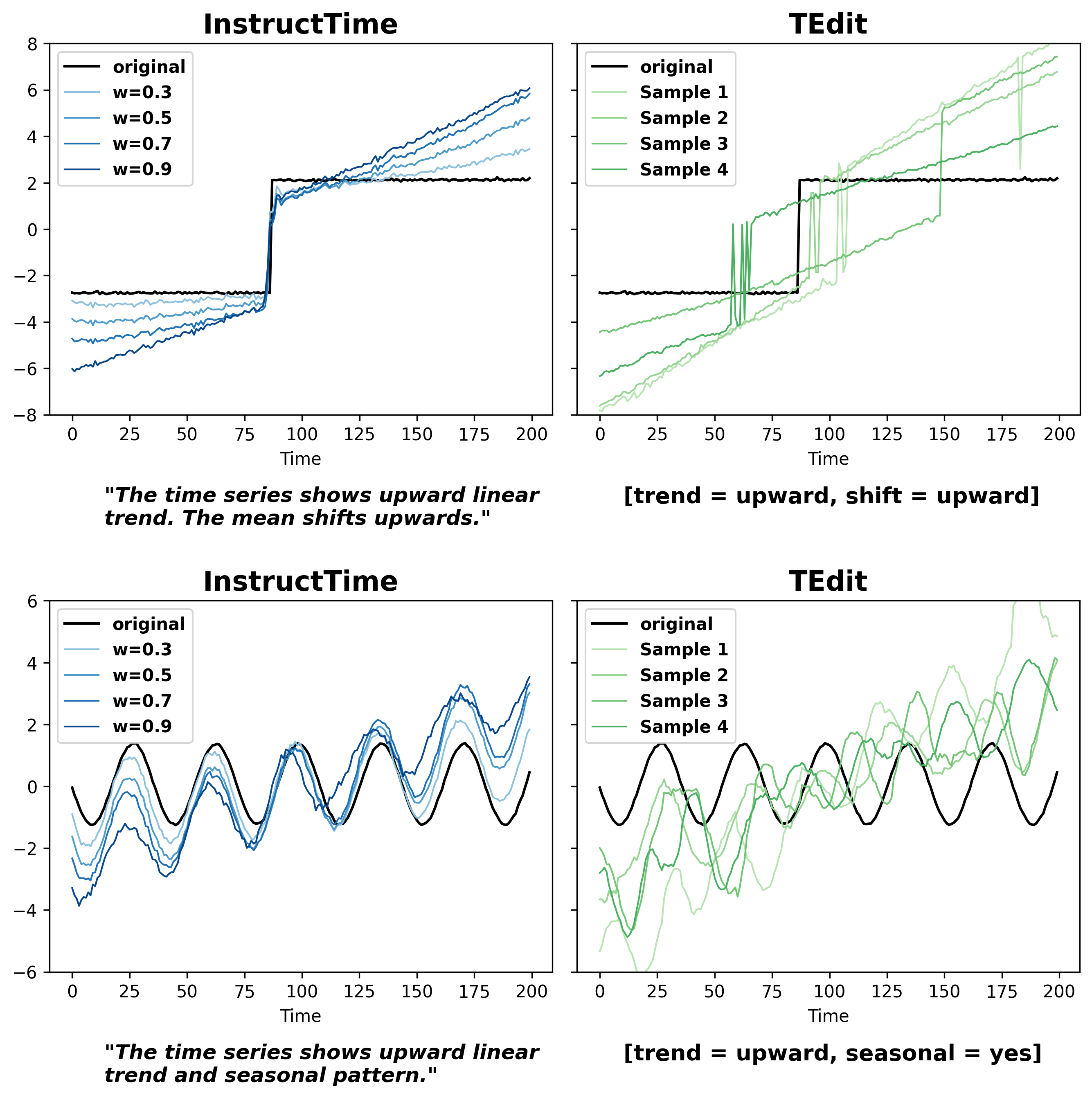}
    \caption{\small Interpolated editing. As the editing strength $w$ increases, InstructTime gradually edits the flat trend into an upward trend while preserving either the local pattern of upward shifts in the mean (upper) or the global pattern of seasonality (lower). Diffusion-based TEdit generates upward trends while preserving other attributes, but with varying slopes and scattered distances from the input.}
    \label{fig:interpolated_case_study}
\end{figure}

\subsection{Editing Quality}
Results for our main comparisons can be seen in Table~\ref{tab:comparison_table}, where we compare each method's editability and preservability for both instruction-based and attribute-based time series editing.
First, we observe that InstructTime is the state-of-the-art editor in the instruction-based setting across all datasets. 
Under the instruction-based setting, InstructTime achieves higher RaTS, MSE, and MAE, along with lower $|\text{RaTS}|$ on the synthetic dataset, compared to alternative editors. 
The higher RaTS indicates that the edited time series more strongly exhibits the target conditions to edit than the input, while the lower $|\text{RaTS}|$ reflects better preservation of the conditions that are not meant to change. The lower $\Delta \text{DTW}$ further shows that the edited time series is closer to the observed ones with the target conditions. The lower MSE and MAE indicate more accurate point-wise alignment between the edits and the synthetic ground truth.
These results suggest that InstructTime more effectively decomposes multiple conditions from an instruction—both those that match the input and those that differ—and edits the time series toward the target conditions while preserving the unique structure of the input instance.
We next note that InstructTime's time series encoder and decoder are agnostic to the natural language instruction encoder. 
Therefore, we also consider the attribute-based setting, which involves no instructions.
We first note that InstructTime is highly-competitive, even outperforming the comparisons in multiple metrics and datasets.
This suggests that our architecture is generalizable and likely drives some of the performance in the instruction-based setting.
We also observe that TEdit shows strong editability and preservability, while InstructTime achieves lower MSE and MAE on synthetic data, indicating better point-wise alignment. 
These results suggest that diffusion-based methods can faithfully edit target attributes, but the output time series are scattered more widely from the input time series due to sampling randomness.
Editing quality on individual attributes can be found in Appendix~\ref{sup:more_results}, alongside edited examples.

\subsection{Interpolated Editing}

\begin{figure*}
    \centering
    \includegraphics[width=\linewidth]{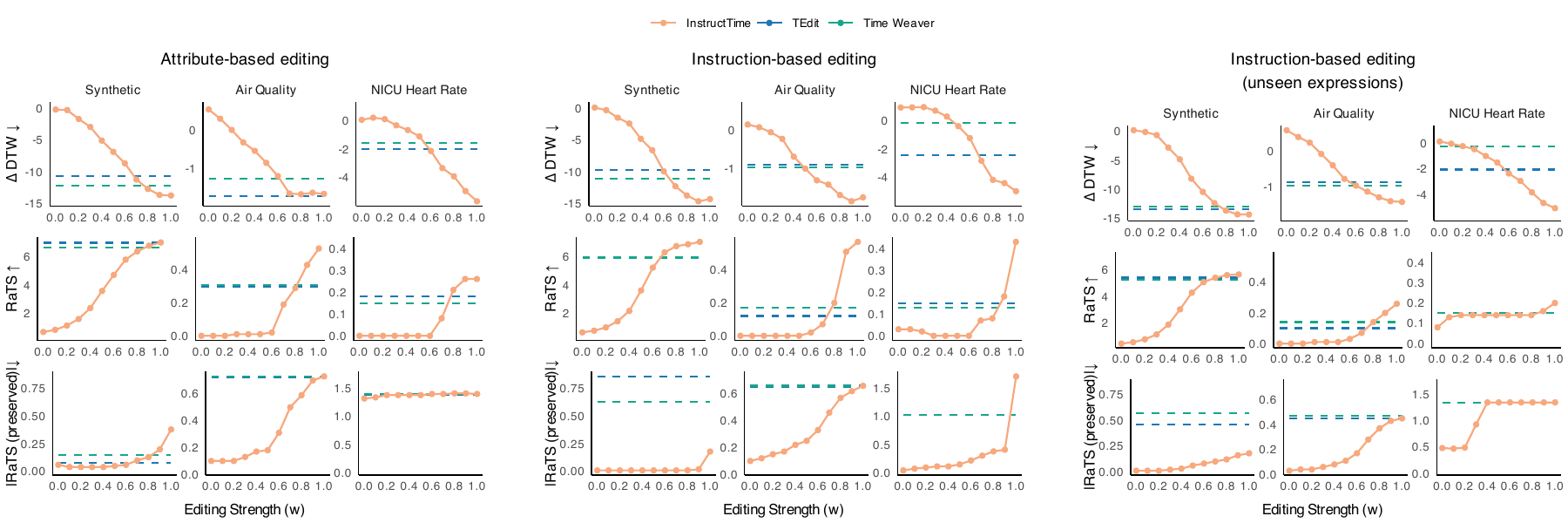}
    \caption{Progressive change in editability and preservability by InstructTime as editing strength $w$ increases from 0 to 1 (step 0.1). TEdit and Time Weaver edit time series without controllable strength. 
    }
    \label{fig:interpolated_editing}
\end{figure*}


\begin{table*}[htbp]
\centering
\resizebox{\textwidth}{!}{%
\begin{tabular}{llccccccccccc}
\toprule
 &  & \multicolumn{5}{c}{\textbf{Synthetic}} & \multicolumn{3}{c}{\textbf{Air quality}} & \multicolumn{3}{c}{\textbf{NICU heart rate}} \\
\cmidrule(lr){3-7}\cmidrule(lr){8-10}\cmidrule(lr){11-13}
 &   & \footnotesize $\Delta$DTW~$\downarrow$& \footnotesize RaTS ↑& \footnotesize |RaTS| ↓& \footnotesize MSE ↓& \footnotesize MAE ↓& \footnotesize $\Delta$DTW~$\downarrow$& \footnotesize RaTS ↑& \footnotesize |RaTS| ↓& \footnotesize $\Delta$DTW~$\downarrow$& \footnotesize RaTS ↑& \footnotesize |RaTS| ↓\\
\midrule
\multirow[c]{3}{*}{\makecell{Instruction-based\\(unseen-phrasing)}} & Time Weaver & -12.95 & 5.24 & 0.57 & 3.25 & 1.48 & -0.97 & 0.14 & 0.47 & -0.27 & 0.15 & \textbf{1.34} \\
 & TEdit & -13.41 & 5.43 & 0.46 & 3.65 & 1.56 & -0.87 & 0.10 & 0.45 & -2.04 & 0.15 & 1.35 \\
 & InstructTime & \textbf{-14.26} & \textbf{5.62} & \textbf{0.16} & \textbf{2.96} & \textbf{1.39} & \textbf{-1.41} & \textbf{0.20} & \textbf{0.43} & \textbf{-4.62} & \textbf{0.16} & 1.35 \\
\bottomrule
\end{tabular}
}
\caption{Editing quality with unseen expressions in instruction-based editing. Scores are averaged across attributes. InstructTime scores are computed at $w = 0.9$.}
\label{tab:unseen_expression}
\end{table*}
As the editing strength $w$ increases, the edited time series should progressively reflect the target edited attributes $\Tilde{\mathbf{c}}_\mathrm{edit}$ to a greater extent, while maintaining high preservability of $\Tilde{\mathbf{c}}_\mathrm{prsv}$.
Figure~\ref{fig:interpolated_case_study} shows a case study comparing the editing procedures by InstructTime and TEdit. As editing strength $w$ increases, InstructTime gradually turns a flat trend into upward trends with increasing slope, while preserving local shifts or global seasonality. In contrast, TEdit produces upward trends with random slopes and scattered distances from the original while preserving other attributes.
Figure~\ref{fig:interpolated_editing} shows the progressive change in editability ($\small \Delta \text{DTW}$, $\small \text{RaTS}$) and preservability ($\small |\text{RaTS}|$) by InstructTime as the editing strength $w$ increases, compared to the fixed performances by TEdit and Time Weaver.  
As $w$ increases, the target attributes become more prominent in the edited time series, resulting in stronger editability. Larger $w$ also leads to greater divergence from the input time series and lower preservability, as changes to target attributes may introduce turbulence in other attributes meant to be preserved. Despite this inherent trade-off, InstructTime maintains comparable preservability as $w$ approaches 1 and achieves better preservability in the low-$w$ regime than other methods.
Compared to attribute-based editing, InstructTime maintains high editing quality in the instruction-based setting, while other methods show reduced editability and preservability, as indicated by higher $\small \Delta \text{DTW}$, lower $\small \text{RaTS}$, and higher $\small |\text{RaTS}|$.

\vspace{-10pt}
\subsection{Generalizability to Unseen Instructions}
\begin{figure}
    \centering
    \includegraphics[width=\linewidth]{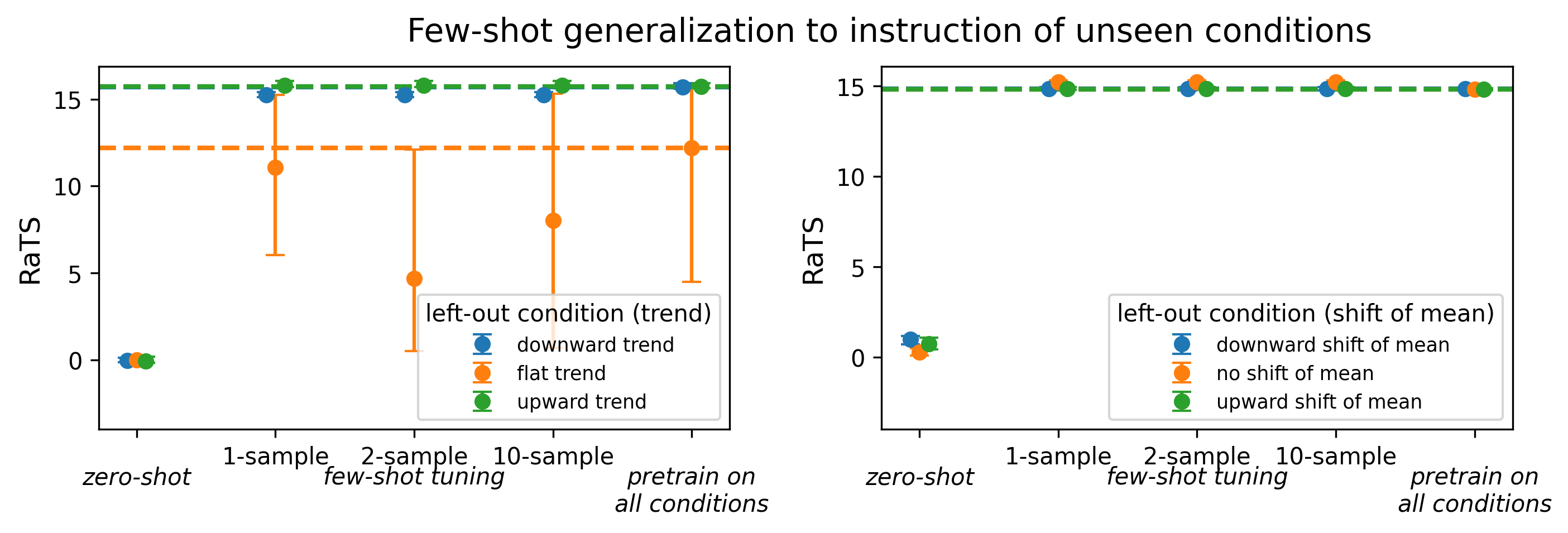}
    \caption{Few-shot tuning InstructTime on unseen conditions approaches the editability by model trained on all levels.}
    \label{fig:few_shot}
\end{figure}

An advantage of editing time series using natural language instructions is the potential to generalize to rich semantics of time series conditions, including (1) new expressions describing known conditions and (2) instructions on entirely unseen conditions. 

For the first case, leveraging a pretrained text encoder allows the instruction encoder to map new expressions of known conditions close to seen ones in the embedding space due to their high cosine similarity. As a result, InstructTime is expected to generalize to such expressions (augmented by GPT-4o) in a zero-shot manner.
As shown in Table~\ref{tab:unseen_expression}, InstructTime achieves comparable editability and preservability when instructions use unseen expressions to describe time series attributes seen during training, and continues to outperform existing methods.

For the second case, we assess InstructTime's leave-one-out generalizability on the 3-level attributes of the synthetic time series: \textit{trend direction} and \textit{shift in mean}. For each attribute, one level (e.g., “upward trend”) is excluded during training. This setup simulates a realistic scenario where the model encounters new but semantically related conditions at inference time. For example, a flat trend can be seen as an average state between upward and downward trends.
We then apply the few-shot tuning procedure using 1, 2, or 10 example instruction–time series pairs from the left-out level. The goal is to test whether limited supervision enables the model to generate edits by leveraging the semantic relationship between unseen and seen conditions.
As shown in Figure~\ref{fig:few_shot}, the model trained on two levels cannot directly generalize to the left-out level in a zero-shot manner, as indicated by the near-zero RaTS scores. However, tuning with a few examples—even just one—results in editability on the unseen level comparable to the baseline where the model is trained on all three levels.

\begin{figure}
    \centering
    \includegraphics[width=\linewidth]{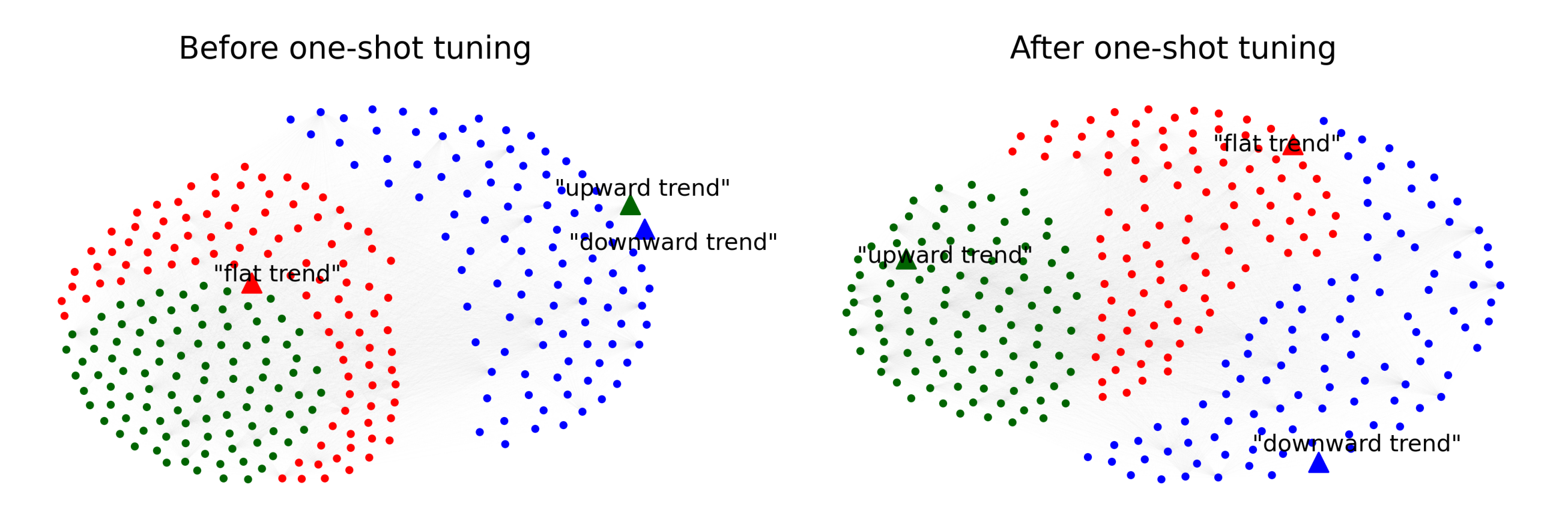}
    \caption{Embedding space before and after tuning with one example from the unseen condition (upward trend). Triangles and circles represent text and time series embeddings, respectively; colors indicate three trend levels. Embeddings are pairwise connected by cosine similarity and visualized in a force-directed layout. }
    \label{fig:one_shot}
\end{figure}
\begin{figure*}
    \centering
    \includegraphics[width=\linewidth]{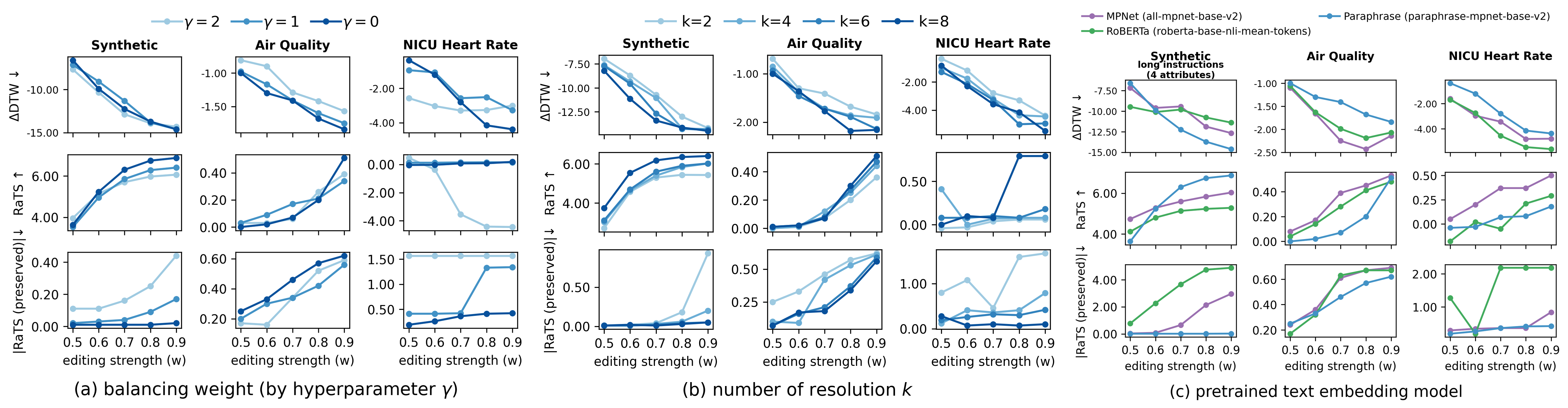}
    \caption{Sensitivity analysis on key components of InstructTime in instruction-based editing.}
    \label{fig:hyper}
\end{figure*}

Figure~\ref{fig:one_shot} visualizes the re-calibration of time series and instruction embeddings in the shared embedding space after tuning with a single example (one-shot). 
Before tuning (left panel), the model is trained on two trend levels: flat (red) and downward (blue). Dots represent embeddings of 100 time series per level; triangles denote their instruction embeddings. The instruction and time series embeddings align well for the seen levels, but for the unseen “upward trend” (green), the triangle and dots are far apart in the embedding space.
The model is then tuned on a small set of instruction–time series pairs, generated by providing an example pair from the unseen level “upward trend” and editing it toward the seen levels to create a few synthetic samples for tuning. This allows the model to adjust its internal representation of the unseen condition in terms of its relative position and distance to the seen conditions in the embedding space. As shown in the right panel, the time series and instruction embeddings are well calibrated across all three levels, highlighting InstructTime’s ability to generalize under limited supervision.

\vspace{-10pt}
\subsection{Ablation Studies}

\subsubsection{Balancing weight.}
When training InstructTime, we jointly optimize a contrastive loss and a reconstruction loss, with their relative contributions to the total loss controlled by a balancing weight. The weight is set so that the unit contribution of the reconstruction loss to the total loss is $10^{-\gamma}$ relative to the contrastive loss. A larger $\gamma$ prompts the model to prioritize the alignment between time series and instruction embeddings, compared to reconstruction.
%
We experiment with $\gamma \in \{0, 1, 2\}$ across all three datasets, corresponding to unit reconstruction-to-contrastive loss ratios of $\{1, 0.1, 0.01\}$, respectively. As shown in Figure~\ref{fig:hyper} (a), both $\gamma = 0$ and $\gamma = 1$ yield similar optimal editability ($\Delta \text{DTW}$, RaTS) and preservability ($|\text{RaTS}|$) across datasets. For the synthetic and NICU heart rate datasets, however, setting $\gamma = 2$ overly de-prioritizes reconstruction, resulting in degraded RaTS and $|\text{RaTS}|$. These suggest while the optimal $\gamma$ may vary by dataset, setting $\mathbf{\gamma \in \{0, 1\}}$ provides a robust and effective default.

\vspace{-5pt}
\subsubsection{Resolution $k$.} 
The multi-resolution time series encoder consists of $k$ CNNs, each with a kernel size set to a different fraction of the input time series length, allowing the model to capture patterns at multiple resolutions. We evaluate the effect of varying $k$ on model performance, with $k = 2, 4, 6, 8$. 
As shown in Figure~\ref{fig:hyper} (b), $\mathbf{k \in \{6,8\}}$ generally yield better editability ($\Delta \text{DTW}$, RaTS) and preservability ($|\text{RaTS}|$) across datasets, whereas $k = 2$ is overly simplistic for capturing complex patterns at varying resolutions for the synthetic and NICU datasets.

\vspace{-5pt}
\subsubsection{Pretrained text embedding model.} 
The instruction encoder leverages a frozen pretrained text embedding model to firstly convert instructions into numerical vectors. The choice of pretrained model should be sensitive to multiple attributes described in a short-paragraph (multi-sentence) instruction to ensure that both the conditions to edit and to preserve are accurately reflected in the edited time series. 
We next experiment with two additional publicly available, computationally efficient pretrained models—MPNet (\textit{all-mpnet-base-v2}) \cite{song2020mpnet} and RoBERTa (\textit{roberta-base-nli-mean-tokens}) \cite{liu2019roberta}—in comparison with our default choice, Paraphrase MPNet (\textit{paraphrase-mpnet-base-v2}) \cite{reimers-2019-sentence-bert}. As shown in Figure~\ref{fig:hyper}(c), all models show improved editability with increasing editing strength. RoBERTa yields the lowest preservability. MPNet performs better on short instructions, while Paraphrase MPNet notably outperforms other models on synthetic data with long instructions describing 4 attributes. This result suggests that MPNet models are effective candidates, as models fine-tuned for paraphrase tasks perform better on long instructions.

\vspace{-8pt}
\section{Conclusion}\label{sec:conclusions}

In this paper, we introduce Instruction-based Time Series Editing, a novel task that enables fine-grained modification of time series based on natural language instructions. Unlike prior attribute-based approaches, our formulation allows flexible conditioning on free-form text, aligning more closely with real-world scenarios where text captures nuanced, individualized context. We propose InstructTime, the first instruction-based time series editor that uses contrastive learning to embed time series and instructions in a shared space, and decodes interpolated embeddings to generate edits under varying degrees of condition influence. 
%
Instruction-based time series editing raises broader considerations, which can drive impactful future research. 
First, foundation models can generate time series using language prompts, but they are not suited for time series editing yet, which requires context-dependent, data-driven models that allow high-fidelity, in-context edits.
Second, fine-grained control over editing strength is essential for instruction-based editors due to the discrete nature of language. Interpolated editing also supports hypothesis generation on how time series transition under varying conditions, but remains exploratory. 
Other future directions may consider: extending the architecture to a variational form~\cite{davidson2018hyperspherical} to enable sampling-based editing with controllable strength and reduced global averaging; enabling step-wise time series editing guided by natural language instructions describing time-dependent conditions; and evaluating real-world utility in domains such as healthcare where time series are accompanied by nuanced and individualized text such as nurse documentation or patient narrative.

\vspace{-8pt}
\section{Acknowledgements}
We thank University of Virginia's High Performance Computing team for computing resources. We thank the reviewers for their constructive feedback, which helps improve the manuscript in many important ways.
Thomas Hartvigsen was supported in part by a CapitalOne faculty fellowship and the University of Virginia's National Security Data and Policy Institute through the U.S. Department of Defense Contracting Activity,	
Award Number \#2024-24070100001. Contact author: Jiaxing Qiu (jq2uw@virginia.edu).

\bibliographystyle{ACM-Reference-Format}
\balance
\bibliography{ref}

\appendix
\balance
\section*{Appendix}


\section{Datasets}\label{sup:datasets}

\subsection{Synthetic Time Series}
\begin{table*}
\centering
\resizebox{0.9\linewidth}{!}{%
\begin{tabular}{llp{6cm}p{9cm}}
\toprule
 Dataset & Attribute & Levels & Templates\\
\midrule
\multirow[t]{4}{*}{Synthetic} & Trend & [no trend, upward linear, downward linear, upward quadratic, downward quadratic] & ``No trend.'' \newline ``The time series shows [upward linear, downward linear, upward quadratic, downward quadratic] trend.'' \\
 & Seasonality & [no, yes] & ``No seasonal pattern.'' \newline ``The time series exhibits a seasonal pattern.'' \\
 & Shift in mean & [no shift, upward shift, downward shift] & ``No sharp shifts.'' \newline ``The mean of the time series shifts [upwards, downwards].'' \\
 & Local variability & [low, high] & ``The time series exhibits [low, high] variability.'' \\
\cline{1-4}
\multirow[t]{2}{*}{Air quality} & City & [Beijing, London] & ``This is air quality in [Beijing, London].'' \\
 & Season & [spring, summer, autumn, winter] & ``The season is [spring, summer, autumn, winter].'' \\
\cline{1-4}
\multirow[t]{2}{*}{NICU heart rate} & Variability & [low, high] & ``The time series exhibits [low, high] variability.'' \\
 & Bradycardia events & [no, yes] & ``[No Bradycardia events. | Bradycardia events happened.]'' \\

\bottomrule
\end{tabular}}
\caption{Natural language templates for each attribute level.}
\label{tab:attr_template}
\end{table*}


We synthesize time series of length $T = 200$ by composing multiple attributes. Each series is formed as
$
\mathbf{x}
= \mathbf{x}_\text{trend}
+ \mathbf{x}_\text{season}
+ \mathbf{x}_\text{shift}
+ \mathbf{x}_\text{noise},
$
where each component is generated according to the attribute configurations below. We consider four multi-resolution attributes: trend type (5 levels: linear upward, linear downward, quadratic upward, quadratic downward, none), seasonality (2 levels: present or absent), abrupt mean shift (3 levels: upward, downward, none), and noise level (2 levels: high or low). For each level of each attribute, we generate 300 time series. Combining components yields $5 \times 2 \times 3 \times 2 = 60$ unique configurations, each with 300 samples, totaling 18{,}000 time series. Trend and seasonality capture global patterns as functions of time, while abrupt mean shifts model local changes at specific time points. We randomly split the dataset into training, validation, and test sets using a 70--20--10 ratio, resulting in 12{,}600 training, 3{,}600 validation, and 1{,}800 test samples.


\noindent\textbf{Trend.}
We generate synthetic time series with five trend types: flat, upward-linear, downward-linear, upward-quadratic, and downward-quadratic. Flat sequences are constant-valued, with each point set to $c \sim \mathcal{U}(-5,5)$. Linear trends follow $x_t = a t$, where $a \sim \mathcal{U}(0.05,0.20)$; $a$ is positive for upward linear and negated for downward linear. Each linear sequence is centered to zero mean and shifted by $m \sim \mathcal{U}(-5,5)$. Quadratic trends use $x_t = a(t+b)^2$, with curvature $a \sim \mathcal{U}(10^{-4},5 \times 10^{-4})$ and shift $b \sim \mathcal{U}(0,50)$; $a$ is positive for upward-quadratic and negated for downward-quadratic. When needed, sequences are shifted upward to ensure non-negativity, then centered and offset by $m$. These constructions allow controlled variation in trend direction and curvature.


\noindent\textbf{Seasonality.}
We generate time series with and without seasonal structure. The \textit{seasonal} group has two equal subsets: one with a \textit{single seasonal component} 
$x_t = A \sin\!\left( \frac{2\pi t}{P} + \phi \right)$, and one with \textit{multiple seasonal components}, formed by summing $k$ such terms with varying amplitudes $A$, periods $P$, and phases $\phi$. All sequences include additive Gaussian noise. This design allows controlled analysis across different levels of seasonal complexity. The \textit{no-seasonality} group uses low-amplitude, long-period oscillations that are effectively indistinguishable from random noise.

\noindent\textbf{Abrupt shift in mean.}
To model localized changes, we simulate series with either no shift, an upward shift, or a downward shift. A \textit{no-shift} series is centered at zero and perturbed by small Gaussian noise $\epsilon_t \sim \mathcal{N}(0,0.05^2)$. For shifted series, an abrupt change occurs at a random time $t_s \in [0.1T,\,0.9T]$, after which all values are offset by $\Delta \sim \mathcal{U}(15,20)$. We apply $+\Delta$ for \textit{upward} and $-\Delta$ for \textit{downward} shifts, producing persistent regime changes.

\noindent\textbf{Noise (local variability).}
We simulate time series with either low or high local variability. Each sequence is composed of additive Gaussian noise $\epsilon_t \sim \mathcal{N}(0,\, \sigma^2)$, where the standard deviation $\sigma$ is drawn from a variability-specific range: $\sigma \sim \mathcal{U}(0.01,\, 0.1)$ for low variability and $\sigma \sim \mathcal{U}(1.5,\, 2.0)$ for high variability.

\subsection{Air Quality}

We combine two public air quality datasets \cite{godahewa2020kddcup, chen2017beijing} previously used in \cite{jing2024towards, narasimhan2024time}, resulting in 3684 time series of hourly PM2.5 from Beijing and London spanning 03/01/2013 to 03/31/2018. The first source \cite{godahewa2020kddcup}, originally used in the KDD Cup 2018, contains hourly air quality measurements from 35 stations in Beijing and 24 in London between January 2017 and March 2018, measuring air pollutants such as PM2.5, PM10, NO\textsubscript{2}, and O\textsubscript{3}. The second source \cite{chen2017beijing} provides hourly PM2.5 data from 12 stations in Beijing from 2013 to 2017. We process the combined dataset following the procedure in \cite{godahewa2020kddcup}: retaining only PM2.5 measurements, slicing each time series into weekly segments of length 168 (24 hours × 7 days), and assigning season labels based on the calendar month. The combined time series are labeled with two attributes: city (2 levels—2,646 from Beijing and 1,038 from London) and season (4 levels—1,169 winter, 932 spring, 839 fall, and 744 summer). Stratified by city, the dataset is randomly split into 2,579 training, 737 validation, and 368 test samples according to a 70-20-10 ratio.

\subsection{NICU Heart Rate}
The NICU heart rate dataset consists of 36{,}679 time series of length 300 (10-minute records sampled every 2 seconds) from 2{,}780 infants at a Neonatal Intensive Care Unit (2012--2016) \cite{nicudataset, sullivan2024comparing}.  
We compute two clinically meaningful attributes for each sequence: heart rate variability (2 levels: low, high) and bradycardia events (2 levels: present or absent). Variability is obtained by dividing each 300-point series into 30 non-overlapping windows of 10 points and taking the median of the window-wise standard deviations. Sequences with values $<\!1.5$ are labeled \textit{low} variability, and those $>\!3$ as \textit{high}. Bradycardia events are identified by detecting contiguous segments where the heart rate drops below 100\,bpm for at most 150 time points; a valid event requires a negative drop rate before onset and a positive recovery afterward.
Of the 36{,}679 sequences, 20{,}000 have low variability and 16{,}679 have high variability; 2{,}147 contain bradycardia events. We split the data into 25{,}675 training, 7{,}336 validation, and 3{,}668 test samples, stratified by variability.


\section{Resolution-wise Semantic Alignment in Contrastive Loss} \label{sup:proof_alignment}

The contrastive loss (symmetric InfoNCE) between a time series embedding $\mathrm{z}_{\mathrm{x}}^{(i)}$ and a instruction embedding $\mathrm{z}_{\mathrm{c}}^{(j)}$ is given as: 

\noindent \resizebox{0.45\textwidth}{!}{$
\mathcal{L}_{\text{contrast}} = \frac{1}{2N} \sum_{i=1}^{N} [
    -\log\left(
        \frac{\exp( \mathrm{z}_{\mathrm{x}}^{(i)} \cdot \mathrm{z}_{\mathrm{c}}^{(i)} )}
             {\sum_{j=1}^{N} \exp( \mathrm{z}_{\mathrm{x}}^{(i)} \cdot \mathrm{z}_{\mathrm{c}}^{(j)} )}
    \right)
    -\log\left(
        \frac{\exp( \mathrm{z}_{\mathrm{c}}^{(i)} \cdot \mathrm{z}_{\mathrm{x}}^{(i)} )}
             {\sum_{j=1}^{N} \exp( \mathrm{z}_{\mathrm{c}}^{(i)} \cdot \mathrm{z}_{\mathrm{x}}^{(j)} )}
    \right)
], 
$}

\noindent where $\mathrm{z}_{\mathrm{x}}^{(i)} \cdot \mathrm{z}_{\mathrm{c}}^{(j)} $ denotes the dot-product similarity between the two embeddings.
Minimizing this loss increases the numerator $\exp(\mathrm{z}{\mathrm{x}}^{(i)} \!\cdot\! \mathrm{z}{\mathrm{c}}^{(i)})$ and decreases the denominator $\sum_{j} \exp(\mathrm{z}{\mathrm{x}}^{(i)} \!\cdot\! \mathrm{z}{\mathrm{c}}^{(j)})$, thereby encouraging each matched pair $(\mathrm{z}{\mathrm{x}}^{(i)}, \mathrm{z}{\mathrm{c}}^{(i)})$ to have higher similarity than any mismatched pair $(\mathrm{z}{\mathrm{x}}^{(i)}, \mathrm{z}{\mathrm{c}}^{(j)})$ for $j \ne i$.

Now, we divide each embedding into $k$ segments:
\[
\mathrm{z}_{\mathrm{x}}^{(i)} = [\mathrm{z}_{\mathrm{x},1}^{(i)}; \ldots; \mathrm{z}_{\mathrm{x},k}^{(i)}],
\quad
\mathrm{z}_{\mathrm{c}}^{(i)} = [\mathrm{z}_{\mathrm{c},1}^{(i)}; \ldots; \mathrm{z}_{\mathrm{c},k}^{(i)}].
\]
The similarity between a pair of embeddings becomes the sum of
their segment-wise similarities:
$\mathrm{z}_{\mathrm{x}}^{(i)} \!\cdot\! \mathrm{z}_{\mathrm{c}}^{(j)}
= \sum_{r=1}^{k} 
\mathrm{z}_{\mathrm{x},r}^{(i)} \!\cdot\! \mathrm{z}_{\mathrm{c},r}^{(j)}.
$
\noindent Then, the first term in the bracket of $\mathcal{L}_{\text{contrast}}$ (symmetry holds for the second term) can be written as: 
\noindent \begin{footnotesize}
\begin{align} \footnotesize
& -\log\!\left(
        \frac{\exp( \mathrm{z}_{\mathrm{x}}^{(i)} \!\cdot\! \mathrm{z}_{\mathrm{c}}^{(i)} )}
             {\sum_{j=1}^{N} \exp( \mathrm{z}_{\mathrm{x}}^{(i)} \!\cdot\! \mathrm{z}_{\mathrm{c}}^{(j)} )}
    \right) = -\log\!\left(
        \frac{\exp\!\left(\sum_{r=1}^{k} 
        \mathrm{z}_{\mathrm{x},r}^{(i)} \!\cdot\! \mathrm{z}_{\mathrm{c},r}^{(i)}\right)}
             {\sum_{j=1}^{N} \exp\!\left(\sum_{r=1}^{k}
        \mathrm{z}_{\mathrm{x},r}^{(i)} \!\cdot\! \mathrm{z}_{\mathrm{c},r}^{(j)}\right)}
    \right) \nonumber \\[4pt]
&= -\log\!\left(
        \frac{\prod_{r=1}^{k} 
        \exp\!\big(\mathrm{z}_{\mathrm{x},r}^{(i)} \!\cdot\! \mathrm{z}_{\mathrm{c},r}^{(i)}\big)}
             {\sum_{j=1}^{N} 
             \prod_{r=1}^{k} 
             \exp\!\big(\mathrm{z}_{\mathrm{x},r}^{(i)} \!\cdot\! \mathrm{z}_{\mathrm{c},r}^{(j)}\big)}
    \right) = -\sum_{r=1}^{k} 
   \log\!\left(
        \frac{\exp\!\big(\mathrm{z}_{\mathrm{x},r}^{(i)} \!\cdot\! \mathrm{z}_{\mathrm{c},r}^{(i)}\big)}
             {\sum_{j=1}^{N} 
             \prod_{r'=1}^{k}
             \exp\!\big(\mathrm{z}_{\mathrm{x},r'}^{(i)} \!\cdot\! \mathrm{z}_{\mathrm{c},r'}^{(j)}\big)}
   \right).  \nonumber 
\end{align}
\end{footnotesize}

\noindent Minimizing these logarithmic terms will increase the segment-wise similarity $\mathrm{z}_{\mathrm{x},r}^{(i)} \!\cdot\! \mathrm{z}_{\mathrm{c},r}^{(i)}$
while suppressing the similarity between mismatched embedding pairs
$\mathrm{z}_{\mathrm{x},r}^{(i)} \!\cdot\! \mathrm{z}_{\mathrm{c},r}^{(j)}$ $(j \ne i)$. 
Therefore, both the embeddings and their corresponding segments become aligned, linking each segment’s semantic representation ($\mathrm{z}_{\mathrm{x},r}^{(i)}$) with its corresponding temporal resolution  ($\mathrm{z}_{\mathrm{c},r}^{(i)}$) during contrastive fusion.

\section{Natural Language Instruction Generation}\label{sup:instruction}



The template for each level of each attribute is provided in Table~\ref{tab:attr_template}. To increase linguistic diversity when evaluating an editor’s generalizability to unseen instructions, we use GPT-4o to paraphrase each distinct attribute level with the following prompt:
\begin{quote}
\texttt{\footnotesize Generate 50 different sentences that have the same meaning as this sentence related to \{context\}: ‘\{template sentence\}’. Only return sentences, do not number the sentences, split the sentences with the character '\textbar' only.}
\end{quote}
Here, \{context\} denotes the dataset domain, and \{template sentence\} is the attribute-level description in Table~\ref{tab:attr_template}. We use temperature 0.1 and split paraphrases 70\%/30\% for training and held-out evaluation.

\section{Additional Experiment Results}\label{sup:more_results}

Across attributes of all three datasets, InstructTime generally achieves optimal editability ($\small \Delta$DTW~$\downarrow$, $\small \text{RaTS} \uparrow$) and preservability ($\small \text{|RaTS|} \downarrow$) under instruction-based settings. A consistent exception is the attribute of local variability (Gaussian-like noise) in the synthetic and NICU datasets, where diffusion-based methods performs better. This suggests that sampling from conditional Gaussian distributions is effective for capturing the varying magnitude of Gaussian noise using prior methods. Incorporating noise estimation mechanisms—such as diffusion models, variational encoder–decoders, or using negative log-likelihood instead of MSE in the loss function—into the InstructTime architecture are important directions for future work.

\end{document}